\documentclass[runningheads]{llncs}

\usepackage{hyperref}
\usepackage{graphicx}
\usepackage{xcolor}
\usepackage[normalem]{ulem}

\newcommand{\add}[1]{\textcolor{black}{#1}}

\begin{document}
	\mainmatter
	
	\title{Process Constitutions and Process Stewards: Towards the Next Generation of BPM for Agentic Organizations}
	\titlerunning{Process Constitutions and Process Stewards}
	
	\author{Amin Jalali\inst{1} \and Majid Rafiei\inst{2}}
	\institute{
		Department of Computer and Systems Sciences, Stockholm University, Sweden,\\
		\email{aj@dsv.su.se},
		\and
		SAP Signavio, Berlin, Germany,\\
		\email{majid.rafiei@sap.com},
	}
	
	\maketitle
	
	\begin{abstract}
		Business Process Management (BPM) was built on a foundational assumption that organizations are populated primarily by human actors whose work can be made visible, governable, and improvable through process models. That assumption is depreciating. AI agent ecosystems increasingly execute, coordinate, and adapt organizational work with limited human direction, challenging not only BPM's methods but its core conception of what a process is. We argue that BPM faces a constitutive shift from modeling human work to governing autonomous agents, for which we propose two new concepts: the \emph{Process Constitution}, a machine-interpretable, value-laden framework that defines the space of admissible agent behavior, and the \emph{Process Steward}, a governance agent that interprets and enforces it. The central value proposition of this new generation of BPM is not efficiency but \emph{organizational legibility}: the capacity to keep agentic organizations accountable, contestable, and humanly understandable. We outline what this means and sketch the research agenda it opens.
	\end{abstract}
	
	
	\section{Introduction}
	
	Business Process Management (BPM) is part of a tradition that seeks to improve how organizations manage their operations through a process-centered perspective~\cite{harmon2010scope}. We abstract its development in terms of two generations. In the first, process models serve as blueprints, providing static specifications of how humans are expected to perform their work. In the second, data and intelligence are layered on top: process logs, mining tools, and predictive analytics turn those blueprints into a dynamic mirror that reflects and optimizes actual behavior~\cite{van2016data}\add{, and, in its AI-augmented form, aims to make execution itself more adaptive and explainable~\cite{dumas2023augmented}}. Both generations share an assumption, namely that humans are the primary organizational actors and that machines automate well-delimited tasks within human-designed structures.
	
	That assumption is fading with the emergence and pervasive use of AI agents capable of executing customer interactions, coordinating work, allocating resources, and in some settings assuming supervisory roles previously reserved for human managers~\cite{andon2024}. 
	Recent empirical evidence also shows how LLM-based systems are increasingly deployed to automate knowledge-intensive business-process tasks that were previously performed by human experts, thereby reshaping process execution and scalability~\cite{khayatbashi2025ai}.
	\add{This challenges not only how processes are managed, but what a process is. BPM has traditionally treated a process as a specifiable, repeatable structure of human-performed activities, fixed in advance by a human designer even as execution varies.} Once \add{agents themselves originate goals and reconfigure their own operating rules in response to context, that structure is no longer fixed by a designer but continuously renegotiated by the agents enacting it: the process becomes an emergent, contested artifact rather than a stable blueprint.}
	
	Recent work on \emph{Agentic Business Process Management} (APM)~\cite{calvanese2026agenticBPM} already argues for an agent-centric extension of BPM in which autonomous, process-aware agents become first-class citizens governed through framed autonomy, explainability, conversational actionability, and self-modification. Against this backdrop, we propose a third generation of BPM in which the central artifact is no longer a blueprint or a mirror but a \emph{constitution}. \add{This third generation complements the AI-augmented BPM systems (ABPMS) vision of Dumas et al.~\cite{dumas2023augmented}: where ABPMS focuses on how a process is executed, our focus is on a related but distinct question, namely which processes and agent behaviors are legitimate to enact in the first place, and who has the authority to decide and contest that.} In this generation, BPM concerns itself with specifying and enforcing \emph{Process Constitutions} that govern ecosystems of human and artificial agents, and with designing \emph{Process Stewards}, specialized governance agents that interpret these constitutions and ensure that emergent agent behavior remains aligned with organizational intent, constraints, and accountability structures.
	
	\section{The Agentic Organization and Its Governance Challenge}
	
	The evolution toward more agentic organizations can be understood as a series of delegation thresholds: first in execution, then in management, and finally in intent formation. Crossing each threshold increases agent autonomy and correspondingly reshapes the role of BPM (cf.~\cite{castelfranchi1998modelling,calvanese2026agenticBPM}).
	
	At the level of \textit{execution}, agents handle operational tasks such as customer service, back-office processing, and logistics routing, acting as performers and generating logs for analysis.
	
	At the level of \textit{management}, organizations increasingly delegate coordination and resource allocation to AI systems that prioritize backlogs, replan projects, and redistribute workloads. Even relatively small operations illustrate this shift: an AI-managed café in Stockholm, for instance, has taken over scheduling and resource-allocation decisions once made by a human manager~\cite{andon2024}.
	When systems of this kind assume process-owner functions, the human process owner is no longer BPM's sole or primary center of governance.
	
	The deepest disruption occurs at the \textit{intent} level. Since processes operationalize intent, delegating both intent and execution to agents does not merely change how processes are enacted; it changes what a process is. Without a new form of governance, BPM risks becoming a thin technical membrane stretched between agentic strategy and agentic operations, with limited influence on either.
	
	Taken together, these thresholds demand a move from workflow specification to constitutional governance.	
	Instead of primarily describing and optimizing sequences of human activities, BPM must define and enforce the high-level structures that constrain, direct, and legitimize the behavior of agent ecosystems. This resonates with the APM vision of framed autonomy, where agent behavior is constrained and aligned through explicit process frames~\cite{calvanese2026agenticBPM}, but we argue that highly agentic organizations additionally require constitutional constructs that define \emph{who} sets those frames and \emph{how} they can be contested and revised.
	
	\section{Process Constitutions, Stewards, and Legibility}
	
	We propose the \emph{Process Constitution} as the core artifact of this third generation of BPM. Unlike BPMN-style models, which prescribe activity sequences and gateway logic, the Process Constitution defines the space of \emph{admissible} processes, that is, the set of processes that agents are permitted to select, specify, and enact. It encodes four interrelated elements: 
	\begin{itemize}
		\item \textbf{Intent:} capturing authorized outcomes, goals, and priorities\add{ -- that is, which outcomes and priorities are legitimate for agents to pursue at all. This is a question of authorship, distinct from an agent's own operational reasoning about how to select, sequence, or prioritize actions in pursuit of a goal that has already been authorized};
		\item \textbf{Constraints:} specifying non-negotiable boundaries such as legal requirements, ethical commitments, safety limits, and sustainability criteria;
		\item \textbf{Accountability:} mapping outcomes to responsible agents and escalation paths; and 
		\item \textbf{Adaptation rules:} specifying when processes and policies may be adapted autonomously and when human authorization is required.
	\end{itemize}

	In APM terminology, constitutions can be seen as a particular, value-laden instantiation of the \emph{normative frames} that constrain agent autonomy~\cite{calvanese2026agenticBPM}, but with a stronger emphasis on authorship, contestability, and explicit allocation of responsibility across human and artificial agents.
	
	\add{We use the term ``constitution'' deliberately, but we are conscious that it has a much longer history outside computer science, in the centuries-long study of constitutional design, the rule of law, and the governance of collective action within political theory, legal philosophy, and institutional economics -- from the social contract and separation of powers~\cite{locke1689two} to legal systems~\cite{hart1961concept}, constitution-making~\cite{elster1995forces}, constitutional political economy~\cite{buchanan1962calculus}, commons governance~\cite{ostrom1990governing,ostrom2005understanding}, and agency relationships~\cite{jensen1976theory}. This paper, being a short position paper grounded in BPM and computer science, does not attempt to import that literature systematically; we flag it because a full treatment of Process Constitutions will need to engage with it directly, in particular on questions of authorship, amendment procedures, and the separation of powers among human and artificial agents.}

	\begin{figure}[t]
		\centering
		\includegraphics[width=1\textwidth]{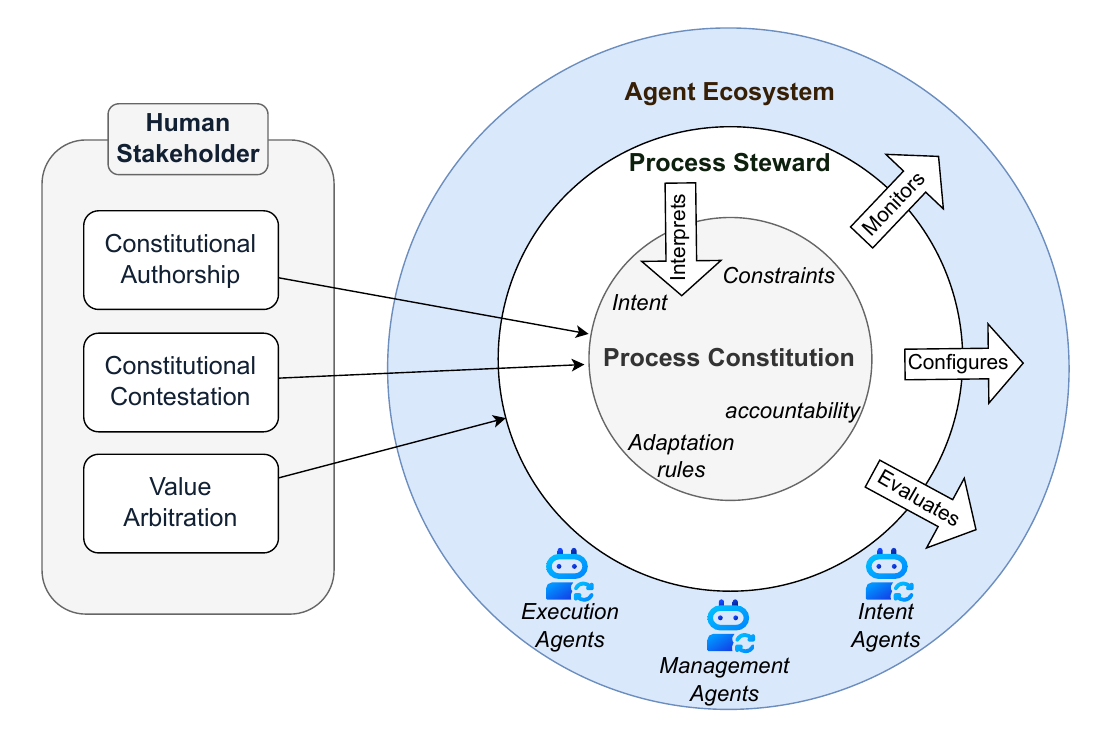}
		\caption{The Process Constitution encodes organizational intent, constraints, accountability, and adaptation rules. The Process Steward interprets the Constitution, then evaluates candidate process designs, configures and allocates agents, and monitors their behavior within the Agent Ecosystem. Human stakeholders hold the sole authority to author and amend the Constitution, override it through value arbitration, and contest Steward interpretations.}
		\label{fig:overview}
	\end{figure}
	
	Figure~\ref{fig:overview} shows an overview of our proposal.
	To become operative, the Process Constitution requires an interpreter and enforcer. We refer to this as the \emph{Process Steward}: a governance agent that evaluates candidate process designs against constitutional constraints, configures and allocates task-execution agents within those limits, monitors behavior for violations, and manages amendments when intents or constraints must change.
	
	\add{This use of ``constitution'' differs from the sense familiar in AI alignment research, where a constitution is a set of principles used to train or prompt a single model's own behavior, most notably in Constitutional AI~\cite{bai2022constitutional}, fixed largely at training or prompting time. The Process Constitution we propose instead governs an organizational ecosystem of many human and artificial agents, is authored and amendable by human stakeholders rather than baked into a model beforehand.}
	
	\add{The Constitution and Steward are not a further, peer-level agent alongside the execution, management, and intent-forming agents introduced in Section~2; they operate one level up, defining and enforcing the boundaries within which those agents may legitimately act. An execution or management agent still decides how to perform or coordinate tasks, and an intent-forming agent still proposes or prioritizes candidate goals; the Constitution instead fixes which outcomes, priorities, and courses of action are admissible in the first place, and the Steward checks that agent behavior at every level stays within that space, rather than performing, coordinating, or originating goals itself.}
	
	The Process Steward is closer to a constitutional court than to a workflow engine: it does not merely orchestrate predefined flows, but interprets high-level commitments, arbitrates tensions among competing objectives, and ensures that emergent patterns of agent behavior remain within constitutionally authorized bounds. 
	
	In the APM architecture, framing, explainability, conversational actionability, and self-modification are distributed capabilities of process-aware agents within a socio-technical system~\cite{calvanese2026agenticBPM}. The Process Steward can be understood as a specialized governance agent that centralizes and reorients parts of these capabilities at the constitutional level: it maintains and applies the frame (the Constitution), explains and justifies governance decisions, engages conversationally with human stakeholders about contested outcomes, and manages controlled self-modification of the Constitution and its own policies.
	
	The deeper purpose of both artifacts is \emph{organizational legibility}. Organizations are social institutions whose legitimacy depends on stakeholders being able to understand, contest, and hold them accountable for what they do. As agent ecosystems grow denser and more autonomous, organizational behavior becomes progressively harder to follow or reconstruct, opening a legitimacy gap. BPM's contribution in this environment may be less about optimizing processes and more about maintaining accountability surfaces that support explanation and contestation. Stakeholders should be able to trace contested outcomes back to specific constitutional elements and to the Process Steward's interpretations.
	
	Even in a fully agentic setting, certain roles remain irreducibly human. \emph{Constitutional authorship}, understood as the articulation of organizational intent, the drawing of ethical limits, and the definition of accountability structures, cannot be delegated to agents without undermining the very idea of constraint. \emph{Constitutional contestation} requires that humans be able to challenge both the Constitution and the Process Steward's interpretations and override them when necessary. \emph{Value arbitration}, the resolution of genuine conflicts among efficiency, fairness, risk tolerance, and other priorities, requires human judgment because such conflicts admit no purely technical resolution.
	
	\section{A Research Agenda}
	
	Taking the Process Constitution and Process Steward seriously opens a set of foundational questions for BPM. While the APM manifesto~\cite{calvanese2026agenticBPM} outlines a rich agenda around framed autonomy, explainability, conversational actionability, and self-modification at the level of agents and socio-technical systems, our focus is on the constitutional layer that governs \emph{which} agent behaviors and process evolutions are legitimate in the first place.
	
	How can organizational intent, constraints, and accountability be represented so that they are simultaneously machine-interpretable and human-auditable? How should a governance agent be architected and evaluated so that it enforces the Constitution without becoming itself opaque or unaccountable? How can organizational legibility be measured, and what thresholds should trigger mandatory human oversight? How can causal and moral responsibility be assigned in outcomes that emerge from interactions among many agents and humans, with no single point of control? And how should interfaces and practices be designed so that humans can perform authorship, contestation, and value arbitration roles without being overwhelmed by the scale and speed of agentic systems? Addressing these questions draws BPM into engagement with AI safety and alignment, organizational sociology, and political theory.
	
	The infrastructures that will shape agentic organizations are being designed now, largely outside the BPM community. The concepts embedded in contemporary agent frameworks, governance platforms, and enterprise AI systems will determine whether agentic organizations remain accountable social actors or become opaque technical systems answerable to no one. BPM has accumulated decades of insight into how organizational work can be made visible, governable, and improvable. If the discipline applies that insight to the constitutional challenges of agentic organizations, rather than restricting itself to the analysis and automation of predefined human-designed processes, it can help ensure that the governance of agentic organizations is something organizations actively design and maintain, rather than something that simply happens to them.
	
	\bibliographystyle{plain}
	\bibliography{references}
	
\end{document}